# Quantum Computational Psychoanalysis
## Quantum logic approach to Bi-logic

Maksim Tomic
Faculty of Sciences, University of Novi Sad, Serbia
[maksaid@live.com](maksaid@live.com)

**Abstract**

In this paper, we are dealing with the fundamental concepts of Bi-logic proposed by Chilean psychoanalyst Ignacio Matte Blanco in the context of quantum logic, founded by Gareth Birkhoff and John Von Neumann. The main purpose of this paper is to present how a quantum-logical model, represented by the lattice of a closed subspace of Hilbert space, can be used as a computational framework for concepts that are originally described by Sigmund Freud as the fundamental properties of the unconscious psyche.

## 1 Introduction

The concepts developed by Chilean psychoanalyst Ignacio Matte Blanco in his books (Matte Blanco, 1980, 1988) represent the first attempt to formalize ideas that Sigmund Freud introduced in his book 'The Interpretation of Dreams' (2011) and then elaborated on his later paper 'The Unconscious'(1915). Matte Blanco's work presents the basis of a scientific discipline that takes a mathematical approach to psychoanalysis- *Computational Psychoanalysis*. Previous work on this topic has mainly dealt with the application of the p-adic model (Khrennikov, 2002), ultrametric topology (Murtagh, 2012; Lauro-Grotto, 2008), and the utilization of grupoid theory to Bi-logic concepts (Iurato, 2014). The author intends to present a new mathematical (logical) model that could formally describe phenomena that represent the fundamental characteristics of the unconscious psyche. In the first part of this paper, we give a short overview of the most important concepts and conclusions of Matte Blanco. The second part of the paper is designed to present the fundamental concepts of quantum logic and to explain the concept of Hilbert space which represents the 'basis' for the quantum-logical system. Eventually, in the third part of the paper, we deal with the interpretation of the phenomena and concepts from Bi-logic in the context of quantum logic.

## 2 Basic concepts of Matte Blanco's Bi-logic

In this section, we address the basic concepts of Matte-Blanco Bi-logic. We give a brief overview of the notions such as the principles of asymmetry, symmetry, generalization, and the implications that follow these principles. A more extensive overview of the basic concepts of Bi-logic can be found in Matte Blanco (1980, 1988), Iurato (2018), Rayner (1995).

First of all, it is necessary to analyze the concepts of the conscious and unconscious psyche that represent the ground for the concepts in Bi-logic. The first one, represented by the principle of **asymmetry**, is very close to a 'classical' logical or analytical thinking whose function is to distinguish different concepts and objects. It is based on definitions (which by themselves make a distinction between notions) and it is a fundamental characteristic of consciousness (Matte Blanco, 1980, p. 37).

Contrarily, the functioning of the unconscious psyche is described by two fundamental principles: **the principle of generalization**} and **the principle of symmetry**. The first of these two is based on the fact that objects, terms belonging to a class describing them (or a set of elements with a particular predicate function), are generalized or represent in the context of the classes that they present. Furthermore, those classes of objects are represented in the context of more general classes, and so on *(*Matte Blanco, p. 38). It should be noted that elements of classes are called objects. The second principle that describes the properties of the unconscious psyche transform, or more precisely treats asymmetrical relations as if they were symmetrical *(*Matte Blanco, p. 38).

We will now outline the essential implications of the principle of symmetry. The first implication is the absence of time (which will be described in more detail) due to the lack of a succession of events. Then, the principle of symmetry implies equality between the whole and its parts, that is, when applied to all the elements that represent one class, we get the equivalence of all members that represent one particular class *(*Matte Blanco, p. 39). Another significant consequence of the principle of symmetry is the equality $p = \neg p$ *(*Matte Blanco, p. 40), which will later be formally described using the propositional calculus of quantum logic.

The aforementioned principles make a clear distinction between how the conscious and unconscious minds operate. We will now list the 5 fundamental characteristics of the unconscious proposed by Freud (Freud, 1915, pp. 186-7), and then explain those characteristics by applying the postulates that had been beforehand explained:

1. The absence of contradiction
2. Displacement
3. Condensation
4. The absence of time
5. Replacement of external reality by physical one

The explanation of listed phenomena:

- The absence of contradiction, or more precisely the identity of a logical variable with its negation, which classical logic does not provide, is explained by the utilization of the principle of symmetry by which all the elements that create a certain class, and so elements $p$ and $\neg p$ are identical.

- Displacement is understood as a tendency of the unconscious psyche to attribute the properties of particular objects to others, that is, to identify one notion with another (either directly or indirectly). In our setting, displacement is directly explained by the principle of generalization by which certain objects are generalized, that is, by their identification (the principle of symmetry). Since all the terms that make up the unconscious psyche can be represented as members of one class, we conclude that it is possible to displace features from any object to every other.

- Condensation is characterized by the combination of the attributes from two or more objects in one object. This feature of the unconscious is also a straightforward consequence of the principle of generalization, and so, of displacement.

- Provided that concept of time in mathematical-physical setting is understood as a successive alternation of events , it is easy to conclude that in case that we have no idea what event happened before (which is due to the principle symmetry), we come to the conclusion that the concept of time does not exist in the unconscious.

- Replacement of physical and external reality (which is one of the fundamental characteristics of the schizophrenic mind) can be explained by the fact that objects that are constituents of the external world, in the unconscious psyche, are identified with the constituents of our psyche, repeatedly because of the principle of symmetry.

## 3 Basic concepts of Hilbert space

This section provides explanations of basic concepts that are essential to understand the concept of a Hilbert space. A more comprehensive summary of the basic concepts of Hilbert space (functional analysis) as well as the basic concepts of linear algebra can be found in Kreyszig (1978), Klein (2013), Sutherland (2010).

Hilbert space represents the mathematical basis of quantum physics and therefore of quantum logic. It is an abstract vector space in which we define our concepts, so it is necessary to determine it.

**Definition 1**. A metric d is a function defined on a set $X$ such that for all x, y, z ∈ X we have:
1. $d \in R^+$
2. $d(x, y) = 0 \Leftrightarrow x = y$
3. $d(x, y) = d(y, x)$
4. $d(x, y) \leq d(x, z) + d(z, y)$

A metric space, with notation (X,d) is an ordered pair of a set with the metric defined on it.

**Defnition 2**. Given a point $x_0 \in X$, and a real number $r > 0$, we define three types of sets:
1. $B(x_0; r) = \{x \in X \mid d(x; x_0) < r\}$ (Open ball)
2. $B'(x_0; r) = \{x \in X \mid d(x; x_0) \leq r\}$ (Closed ball)
3. $S(x_0; r) = \{x \in X \mid d(x; x_0) = r\}$ (Sphere)

**Definiton 3**. A subset M of a metric space X is said to be **open** if it contains a sphere around each of its points. A subset K is said to be **closed** if its complement is open, that is $K^c = X - K$ is open.

**Definition 4**. A sequence $x_n$ in a metric space $X = (X, d)$ is said to be convergent if there is a point $x \in X$ such that:

$$\lim_{n\to\infty} d(x_n, x) = 0$$

**Definition 5.** Let $x = (x_1, x_2, ..., x_n)$ be a sequence in a metric space $X = (X, d)$. This sequence is said to be **Cauchy** if:

$$(\forall \varepsilon > 0)(\exists n_0 \in N)(\forall m, n \geq n_0)(d(x_m, x_n) < \varepsilon)$$

**Definition 6.** The space $X$ is said to be **complete** if every Cauchy sequence in $X$ converges (that is, has a limit which is an element of $X$).

**Definition 7.** A vector space $V$ over $K$, where $K$ is the field of complex or real numbers, is a nonempty set $X$ such that for every element $v \in X$ and a scalar $\alpha$ holds:
1. $V$ contains zero vector
2. $(\forall v \in V)(v \in V) \Rightarrow (\alpha v \in V), \alpha \in R$
3. $(\forall v, z)(v, z \in V) \Rightarrow (v + z \in V)$

**Definition 8.** A subspace of a vector space $X$ is a nonempty set $Y$ which itself contains the properties of a vector space.

**Definiton 9.** A **linear combinations** of vectors $x_1, x_2, ..., x_n$ is an expression of the form:

$$\alpha_1 x_1 + \alpha_2 x_2 + ... + \alpha_n x_n$$

where the coefficients $\alpha_1, ..., \alpha_n$ are any scalars.

**Definiton 10.** The linear closure of a set of vectors $\{v_1, v_2, ..., v_n\}$, written $\text{Span}\{v_1, v_2, ..., v_n\}$, is a set of all linear combinations of that set of vectors.

**Definition 11.** If the expression:

$$0 = \alpha_1 x_1 + \alpha_2 x_2 + ... + \alpha_n x_n$$

can be written in that form iff $\alpha_1 = \alpha_2 = ... = \alpha_n = 0$ then the set of vectors $x_1, x_2, ..., x_n$ is said to be **linearly independent**. Otherwise, this set of vectors is said to be **linearly dependent**.

**Definition 12.** Let $V$ be a vector space, and $v$ any element of it. If vector $v$ could be represented as a linear combination of a set of vectors $\{v_1, v_2, ..., v_n\}$, where this set is linearly independent, then this set of vectors is said to be **basis** of a vector space $V$.

**Definition 13.** Vector space with finite-many vectors in its basis is said to be **finite dimensional**. Otherwise, a vector space with infinitely many vectors in its basis is said to be **infinite dimensional**.

**Definition 14.** A norm defined on a vector space $V$ is a real valued function $\| \ \|: V \to R$ such that for all $x \in V$ satisfied the following properties:

1. $\|x\| \geq 0, \forall x \in V$
2. $\|x\| = 0 \Leftrightarrow x = 0$
3. $\|\lambda x\| = |\lambda| \|x\|, \forall \lambda \in R, \forall x \in V$
4. $\|x + y\| \leq \|x\| + \|y\|, \forall x, y \in V$

A **normed vector space**, in notation $(V, \|\ \|)$, is defined as an ordered pair of a vector space $V$, and a norm defined on it.

**Definition 15**. Let $V$ be a vector space over a field of complex numbers. An **inner product** on $V$ is a mapping $(,): V \times V \to C$ such that for every pair of vectors $x, y \in V$ satisfies the following properties:
1. $(x, x) \geq 0$
2. $(x, x) = 0 \Leftrightarrow x = 0$
3. $(x, z) + (y, z) = (x + y, z)$
4. $(\alpha x, y) = \alpha(x, y), \forall \alpha \in C$
5. $(x, y) = \overline{(y, x)}$

An **unitary vector space**, in notation $(V, (,))$, is represented by an ordered pair of a vector space $V$ with an inner product defined on $V$.

**Definition 16**. A vector $x$ that is the element of a vector space $X$ is said to be orthogonal to $y \in X$ if:
$$(x, y) = 0$$

**Definition 17**. Let $W$ be a vector space over the field of real numbers, and $U$ is a subspace of $W$. The orthogonal complement of $U$ concerning $W$ is a space $V$ such that:
$$V = \{w \in W \mid (v, w) = 0, \forall v \in U\}$$

**Definition 18**. A **Hilbert space**, with notation $H$, is a unitary, complete vector space.

## 4 Basic concepts of quantum logic

In this section, we present the concepts of quantum logic founded in 1936 by Birkhoff and Von Neumann in their work 'The logic of quantum mechanics' (Birkhoff & Neumann, 1936). An overview and interpretation of the concepts of quantum disjunction, conjunction or negation are given. Further on this subject can be found in Pavičić, Megill (2009), Svozil (1999), Rédei (1998), Holik et al. (2012), Khrennikov (2015).

Officially, the term *quantum logic* was first mentioned in 1936 by Birkoff and von Neumann in their work "The Logic of Quantum Mechanics". The essential idea of quantum logic is that certain properties from quantum mechanics are represented by closed vector subspace, more precisely the lattice of closed subspace (projections) of Hilbert space (Redei, 1998; Holik et al., 2012; Holik, 2014). We will now outline some basic algebraic properties of such a logical system.

The first algebraic structure that Birkoff and von Neumann thought would represent the algebraic basis for quantum logic is ortholattice $<L(H), \wedge, \vee, ', 1, 0>$. The main problem that they identified in such a structure is that distributivity is lost:

$$x \wedge (y \vee z) \neq (x \wedge y) \vee (x \wedge z)$$

The structure which they then identified as more suitable, is a modular lattice, which for finite-dimensional Hilbert spaces satisfies the *modular* law :

$$x \leq y \Rightarrow x \vee (y \wedge z) \Leftrightarrow y \wedge (x \vee z)$$

However, in the case of an infinite-dimensional Hilbert space, the modular law does not endure. One year after the publication of their work, Kodi Husimi in 1937 (Foulis et al., 2016) showed that in this case the 'weaker' variant of a modular law, the so-called *orthomodular* law, is maintained:

$$x \leq y \Rightarrow x \vee (x' \wedge y) \Leftrightarrow y$$

Then, in the end, an orthomodular lattice (Pavičić & Megill, 2009) opposite to Boolean, which represent the basis for Boolean logic, was taken as the algebraic model for quantum logic.

The central point that makes the discrepancy between quantum and Boolean logic is its interpretation. In the case of quantum logic, the logical variable is seen as a closed subspace of a Hilbert spaceThe explanation of basic logical operations is the following:

- In the case of conjunction (Redei, 1998, p.46; Khrennikov, 2015) it represents the intersection of two closed subspaces. As for the truth value of quantum conjunction, if the vector $v$ belongs to the subspace $A$ ( $A$ has a true valuation), and also belongs to the subspace $B$ ( $B$ also has a true valuation), then vector $v$ will also belong to their intersection, and then the conjunction has the valuation true. Oppositely, the vector $v$ does not belong to their intersection, and then the conjunction is false. We will now represent a more formal explanation of quantum conjunction. If $p$ and $q$ are two logical variables and $M_p$ and $M_q$ are closed subspaces corresponding to them, then the quantum conjunction is represented as:

$$M_{p \wedge q} = \{x \mid x \in M_p \wedge x \in M_q\}$$

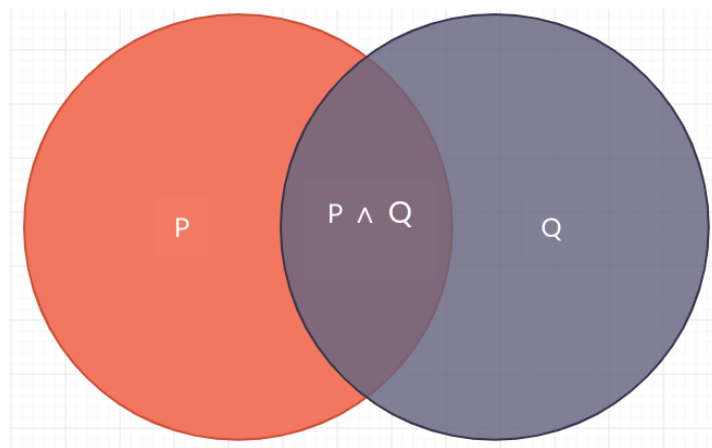

Figure 1: Ilustration of quantum conjunction

- In the case of negation (Khrennikov, 2015), we consider the orthogonal complement of the subspace $A$. Given an arbitrary logical variable $p$ and a subspace associated with it, $M_p$, then the negation in quantum logic can be written as:

$$M_{p'} = \{x \mid (x, y) = 0, y \in M_p\}$$

If the variable $p$ is true, then its negation will be false, as in the opposite case. However, if the vector $v$ is not in the subspace $A$, this does not necessarily imply that it will be in the orthogonal complement of the subspace $A$, which would mean that the formula $p \vee p'$ is not always a tautology.

- The disjunction in quantum logic (Aerts et al., 2000; Khrennikov, 2015), in the notation $\vee$, is interpreted as the linear closure of two subspaces of Hilbert space. Thus for the logical variables $p$ and $q$ and their subspaces $M_p$, $M_q$ respectively, the quantum disjunction is formally expressed as:

$$M_{p \vee q} = \{x \mid x = \alpha y + \beta z, \alpha, \beta \in C, y \in M_p, z \in M_q\}$$

This would mean that if the vector $v$ belongs to space $A$ or (and) space $B$, it will also be in their linear closure. However, if the vector $v$ does not lie in any of these two subspaces, this does not imply that it does not lie in their linear closure, for the plain reason that linear closure involves all possible linear combinations of vectors that make up the given space. This would mean that the disjunction of the two terms may be true even if both logical variables are false.

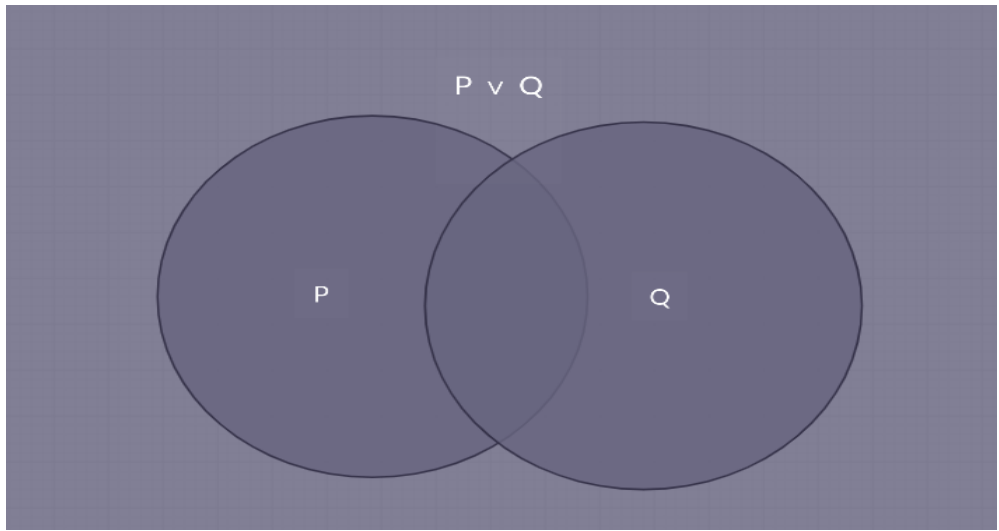

Figure 2: Ilustration of quantum disjunction

# 5 Quantum logic approach to Bi-logic

In this section, we will try to give an interpretation of the earlier described Bi-logic phenomena in the context of quantum logic.

## 5.1 Fundamental approach

We will now try to present the specified concepts of Bi-logic more formally. Let the variable $p$ be observed as a conjunction of the variables $x_1, x_2, ..., x_n$. In the context of quantum logic, the representation of a variable $p$ is, therefore, a closed subspace that will represent the intersection of the subspaces $X_1, X_2, ..., X_n$, while from Bi-logical point of view it is viewed as an object found in the classes of objects represented by those closed subspaces. In this way, we have uniquely identified each variable or object, which, for example, would have a semantic interpretation 'a professor who is married and works at a university'. In this case, the statement variable $p$ could be expressed as follows:

$$p = x_1 \wedge x_2 \wedge x_3$$

where $x_1$ represents a class of objects that have the attribute 'professor' (that is, the set of elements with the specified predicate function), $x_2$ represents a class of people who work at a university, while $x_3$ represents those objects that possess the attribute 'married'. It should be noted here that the aforementioned closed subspaces $X_1, X_2, ..., X_n$ are spaces of the same dimensions.

## 5.2 Principle of generalisation and principle of symmetry in the context of quantum logic

We will now take a look at the concept of generalization in the context of quantum logic. Namely, generalization as one of the basic characteristics of the unconscious can be expressed in terms of a quantum disjunction, that is, a linear closure of subspaces that were used to define a particular object (that is a statement variable). In this way, a certain object is generalized. Now we will return to the illustrated example of a professor who is married and works at the university. If we wanted to 'generalize' the variable $p$, we would then represent it as:

$$p = x_1 \vee x_2 \vee x_3$$

In that case, we would get a definition that generalizes the variable $p$. Using this representation, we would encompass all objects that are constituents of the classes $x_1, x_2, x_3$, or, all the elements that make up their corresponding $X_1, X_2, X_3$. What is crucial to perceive here is that in this case, the variable $\neg p$ could also be represented in the same form (since the orthogonal complement belongs to the same linear closure), precisely:

$$\neg p = x_1 \vee x_2 \vee x_3$$

From this, it follows the conclusion:

$$\neg p = p$$

We have illustrated here the effectiveness of a quantum-logical notation in explaining the concepts of Bi-logic. Now, it is important to pay attention to the concept of 'quantum logical indeterminacy'. This concept is based on the fact that in the 'general' representation of logical variables (the representation in which quantum disjunction is used) is indeterminate how that variable looks like. More concretely, there are **more** options for what a logical variable would look like in that case. In this way the essence of the process of generalization in the unconscious is depicted, because we are able to interpret in various ways what the term $p = x_1 \vee x_2 \vee x_3$ actually represent, because of the definition of a linear closure itself, which includes **all possible linear combinations**. This means exactly that we could represent a whole space (class of objects) using a single propositional variable in its generalized notation. On the other hand, this approach brought us to the idea that Matte Blanco has labeled as the principle of symmetry, that is precisely one of its most important implication, the identification of all members from one class. We will now demonstrate this implication through another example. Let the logical variables $p_1, p_2, p_3$ represent objects with semantic interpretation 'married professors that work at the university' (we will remain in the same example). Then their generalized representation is:

$$p_1 = x_1 \vee x_2 \vee x_3$$
$$p_2 = x_1 \vee x_2 \vee x_3$$
$$p_3 = x_1 \vee x_2 \vee x_3$$

From this it follows:

$$p_1 = p_2 = p_3$$

From the example above, we see that due to the application of the principle of generalization, all objects (more precisely all in the aforementioned example, which is certainly the case an arbitrary number of objects) that belong to the same set of classes are equalized (symmetry principle). Lastly, following this procedure, it is conceivable to obtain equality between whole and its parts (because in this way each statement variable is represented by an entire subspace), which is also an significant implication of the principle of symmetry.

**5.3 Condensation in the context of quantum logic**

Condensation, that is, the principle by which the properties of a particular object of a class are attributed to another object of the same class, could also be understood in the context of quantum disjunction, that is, the 'linear combination' of features of two (or more) objects. Take for example an object that represents a blue-eyed person that works in a government. Let it be represented by a statement variable $r$ and corresponds to the subspace $X_r$. Logical variable $r$ could then be written in the form:

$$r = x_1 \wedge x_2$$

where $x_1$ represents the class of objects that possess the attribute 'blue-eyed person', while $x_2$ represents the class of objects that have the attribute 'works in a government'. Now let us define the variable $q$, which represents an object with the feature 'person who has green eyes and works as a doctor'. Therefore, it corresponds to the subspace $X_q$, and we formally represent it as:

$$q = x_3 \wedge x_4$$

where $x_3$ represents the class of objects with the characteristic 'posses green eyes' and $x_4$ correspond to the class of those objects with the attribute 'doctor'. If we wanted to utilize the condensation principle to these two objects, that would give a new object $p$ with the properties that are a mixture of properties that exist in $q$ and $r$, and we could formally represent that new object as:
$$p = q \vee r$$
In this way, an object that is represented by a variable $p$ would have the properties of both objects, that is, it could represent 'a person with an eyeshade between blue and green, who is a doctor and works in the government'. Using this method we can produce an infinite number of linear combinations of the properties of individual objects. Hence, the object takes on a large number of various properties, which were the properties of other objects.

### 5.4 Displacement in the context of quantum logic

The difference between the formal approach to displacement and condensation is almost imperceptible. Semantically, the two concepts differ in that when displacement occurs, individual attributes of a particular object are attributed to other objects (thus, these objects begin to resemble each other to a greater or lesser extent), while in the condensation properties of two or more objects combine forming a new one. Let $f$ be a logical variable that represents an object with the attribute ' person with two children and a university diploma'. Then, let $h$ be a variable that represents an object with the attribute 'person who is unemployed and has blue eyes'. Formal representation of the described statement variables is:
$$f = x_1 \wedge x_2$$
$$h = x_3 \wedge x_4$$
where $x_1, x_2, x_3, x_4$ are classes of objects that have the attribute: 'has two children', 'has a university degree', 'is unemployed', 'has blue eyes', respectively. Displacement would then give us an object, represented by variable $f$, that, for example, has the feature 'person with two children, who has blue eyes and a university diploma'. Following this procedure the variable $f$ would be represented as:
$$f = (x_1 \wedge x_2) \vee (x_3 \wedge x_4)$$
From a logical point of view, condensation, as noted above, would represent a classical linear closure of two or more subspaces, while displacement could be delineated by the fact not necessary all the vectors that constituent a base of a space (an object that 'transfers' its properties to another) can be used, that is, they can cancel. In the example we have given, this would mean that the property 'unemployed' from object represented by a variable $h$ does not have to be 'passed' to an object represented by a variable $f$. It should be noted here that it is possible to achieve various combinations of properties that an object represented by a variable $f$ may have, that is, it may be absolutely identical to an object represented by a variable $h$ (retaining solely its properties), does not change at all, or take just a meager part of the properties from another object.

### 5.5 Quantum logical mode – final remarks

In the final remarks regarding the quantum-logical model, we have to mention several things. First of all, we have to explain the remaining two properties of the Unconscious psyche which are the absence of time and replacement of the external reality by the internal one. The simplest way to represent these two aspects is to represent them as another property, which is a feature of some particular object, and that operates with them in the same way we operate with the 'regular' attributes. To be more specific, we could add a temporal component to some event describing it with 'happened a year ago', 'last week', 'a couple of days ago'. The situation is a little bit more complicated with the spatial properties because we will have to add some specific functions of the self to define some event as 'real' or 'imaginary'. Adding these, final remarks to the quantum-logical model explained in this, final, section, we believe that we have achieved the task proposed in the introductory section of this article which is representing a formal (quantum-logical) way to describe essential concepts from Matte-Blanco's Bi-logic.

# 6 Conclusion

This paper should mainly serve as a base for further study of a given topic, that is, primarily as a basis for applying a quantum-logical model to formalize the concepts of Bi-logic. We demonstrated that the propositional calculus of quantum logic (lattice of closed subspace) can be used to formally explain the basic principles of Bi-logic introduced by Matte-Blanco. Furthermore, we pointed out the possibility of application of quantum logic to the field of Cognitive Science in general. In the first 3 sections we introduced a theoretical framework for the application of quantum logic to the field of Computational Psychoanalysis, and explain that application in the final section of the paper. The conclusions drawn in the last section of the text should not be understood as ultimate but as the first effort to formalize the fundamental characteristics of the unconscious from the quantum-logical point of view. In the end, it is worth mentioning that, in this paper, we have used the basic formalisms of quantum logic, which certainly could be extended by others in the context of functional analysis and quantum logic.

**Acknowledgmenets**
I would like to thank professors Fionn Murtagh and Andrei Khrennikov, as well as Giuseppe Iurato for their expert guidance, and their willingness to consider my ideas.